# Open Set Recognition for Random Forest

### Guanchao Feng
guanchao.feng@blackrock.com
BlackRock, Inc.
New York, NY, USA

### Stefano Pasquali
stefano.pasquali@blackrock.com
BlackRock, Inc.
New York, NY, USA

### Dhruv Desai
dhruv.desai1@blackrock.com
BlackRock, Inc.
New York, NY, USA

### Dhagash Mehta
dhagash.mehta@blackrock.com
BlackRock, Inc.
New York, NY, USA

## ABSTRACT

In many real-world classification or recognition tasks, it is often difficult to collect training examples that exhaust all possible classes due to, for example, incomplete knowledge during training or ever changing regimes. Therefore, samples from unknown/novel classes may be encountered in testing/deployment. In such scenarios, the classifiers should be able to i) perform classification on known classes, and at the same time, ii) identify samples from unknown classes. This is known as open-set recognition. Although random forest has been an extremely successful framework as a general-purpose classification (and regression) method, in practice, it usually operates under the closed-set assumption and is not able to identify samples from new classes when run out of the box. In this work, we propose a novel approach to enabling open-set recognition capability for random forest classifiers by incorporating distance metric learning and distance-based open-set recognition. The proposed method is validated on both synthetic and real-world datasets. The experimental results indicate that the proposed approach outperforms state-of-the-art distance-based open-set recognition methods.

## 1 INTRODUCTION

Machine learning (ML)–based pattern recognition and classification algorithms have seen unprecedented success in a wide range of applications in different fields such as healthcare, bio-medicine, e-commerce and finance, since they open up many opportunities and improvements to derive deeper and more practical insights from data that can help businesses make informed decisions. However, many conventional ML methods for classification operate under the closed-set assumption, i.e., training set and testing set share the same label space, and are not capable of handling samples from unknown classes (whose samples are unobserved during training). In other words, closed-set settings assume all possible classes are encountered during training. In practice, the closed-set assumption is not only difficult to test but also often violated due to incomplete knowledge at training time since it is often difficult or even impossible to collect training samples to exhaust all possible classes in training in many real-world applications.

On the contrary, open-set setups allow the existence of samples from unknown classes during testing, which are better suited for real-world classification tasks. In the open-set settings, classifiers are required to not only accurately classify new instances of known classes (whose samples are observed during training) but also effectively recognize the samples from unknown classes. In a nutshell, open-set classifiers are capable of making the "none of the above" decision with respect to known classes. This is known as open-set recognition (OSR) [38] and has received significant attention in recent years [11, 47]. Since many learning tasks in finance are naturally classification tasks, for instance, company classifications using Global Industry Classification Standard (GICS), fund categorization, risk profiling, economic scenario classifications, etc., where often a new company, fund or economic scenario may not belong to any of the existing categories, casting these recognition tasks as OSR instead of traditional closed-set classification tasks is more appropriate. For instance, in [14] (see also [20]), the authors investigated miscategorization of funds and showed that a few funds in the data might not belong to any of the existing Morningstar [1] categories [30] according to their outlier measures.

Random forest (RF) introduced in [6] has been successfully applied in a variety of machine learning tasks [5]. The base learners in RF are decision trees constructed using the Classification and Regression Tree (CART)-split criterion [7, 24] whose outputs are then aggregated into a single output using majority voting for classification problems or averaging for regression problems. RF applies randomization to both features and samples for decorrelating trees so that the variance of ensemble is reduced [12]: the data points in training set are sampled with replacement (known as bootstrap aggregation or bagging); at each node, a certain number of features (a randomly selected subset) are adopted [36]. The RF has a number of merits such as robustness to feature scaling, ability to effortlessly handling missing values and outliers [15], inherent ability to handle non-linearity [1], and more importantly, excellent applicability with small sample sizes and high-dimensional feature spaces [5, 51] which are often encountered in tabular data. In fact, tree-based models such as RF often outperform deep learning approaches on learning tasks with tabular data [19, 41]. Tabular data commonly and naturally exists in finance to represent and store information in applications such as fund holdings, exposure to different risk factors, trading, investment management and loan activities. RF models (as well as other tree-based approaches) are widely used in machine learning tasks in finance [27].

Similar to many conventional ML methods for classification tasks, plain-vanilla RF classifiers operate under the closed-set assumption. RF classifiers designed with OSR capability are relative sparse in the literature. In [48], the authors addressed OSR task by combining the extreme value theory (EVT) into RF classifiers in the context of radar high range resolution profile recognition. Particularly, for each known class, a Weibull distribution is fitted independently using extreme value points identified by ranking the Euclidean distance between class members and the class centroid. Given a test sample, its probabilities of belonging to the known classes are derived using the Weibull cumulative distribution, and then the rejection probability (i.e., the probability of belonging to unknown classes) is indirectly calculated through these probabilities. More recently, in [2], for detecting novel traffic scenarios, OSR was implemented by fitting a Weibull distribution for each known class on the number of trees voting for the class. Similar to [48], the rejection probability is then derived accordingly. It is worth noting that, in these setups, the number of Weibull distributions to fit increases with the number of known classes. If any of these Weibull distributions are not well-fitted/estimated, e.g., due to data scarcity, distortions may be introduced in the estimation of rejection probabilities.

In this paper, we propose a novel approach to enabling OSR capability for RF classifiers using Random Forest-Geometry- and Accuracy-Preserving (RF-GAP) proximities and distance-based OSR. Particularly, we first infer a task-specific distance metric from RF-GAP using Gaussian processes (GPs) which are Bayesian machinery for modeling functions. Then the learned distance metric is unitized to perform distance-based OSR. We show that the distance metric inferred from RF-GAP proximities can boost OSR performance. In identification of extreme values, a global EVT model is trained instead of a collection of class-wise EVT models, which is more robust. The proposed approach is first tested on a synthetic dataset generated using a Gaussian mixture model, then tested on popular real-world machine learning benchmarking datasets as well as Morningstar's fund data. The remaining paper is organized as follows. In Section 2, we provide a brief background on OSR, RF-GAP proximity and GP framework. In Section 3, we describe the proposed approach in detail. In the following section, we present experimental results on both synthetic and real-world datasets. Finally, in Section 5 we conclude the paper with some final remarks.

## 2 BACKGROUND

### 2.1 Open-set Recognition

The *open space* is all the region of the feature space outside the support of the training samples. For binary classification, for a given class, its *positively labeled open space* (PLOS) is the intersection of the open space and the positively labeled region which is the region of the feature space in which a sample would be classified as positive. Similarly, in multiclass classification, for all known classes, one can define *known labeled open space* (KLOS) as the union of the PLOS of each class. In short, KLOS is comprised of all the areas of the feature space that lie outside the support of the training samples, in which a sample would be classified as belonging to one of the known classes. Open space risk is considered to be the fraction (in terms of the Lebesgue measure) of KLOS compared to the overall measure of known labeled space, i.e., the region of the feature space in which a sample would be classified as one of the known classes [29, 38].

In closed-set scenarios, the classifiers are trained by optimizing over empirical risk or training error $R_\varepsilon$ measured on training data, whereas in open-set setups, the classifiers should additionally take open space risk $R_o$ (i.e., the unknown) into account. Obviously, there is a trade-off between empirical risk $R_\varepsilon$ and open space risk $R_o$, and the goal of OSR is to achieve a balance between $R_\varepsilon$ and $R_o$. Let $f_r$ represent the learned recognition function of a classifier, then the OSR problem can be formalized as follows:

$$\underset{f_r}{\mathrm{argmin}}\{R_o(f_r) + \lambda R_\varepsilon(f_r)\}, \quad (1)$$

where $\lambda$ is a regularization constant [38]. The equation in (1) provides theoretical guidance as well as insights to OSR modeling and generates a series of OSR algorithms which can be mainly categorized into discriminative models (see, e.g., [16, 32, 46]), and generative models (see, e.g., adversarial learning-based methods [31, 53]). Intuitively, open space risk exists when a recognition model labels feature space far from any training data. From this perspective, naturally, one can address OSR task by limiting distance between test points to training points characterized using e.g., nearest neighbors [29, 42] or nearest centroid [3]. This subset of discriminative approaches are known as distance-based OSR.

The OSR problem is inherently connected with research topics that require proper handling of unseen situations, e.g., lifelong learning [23], transfer learning [44], zero-shot or few-shot learning [34, 37, 43, 49], outlier/novelty detection [33], out-of-distribution detection [18, 52] and so forth. In [3], the authors generalized the OSR to open-world recognition which further requires the model to track the encountered samples from unknown classes, then incrementally learn and extend the multi-class classifier. In this work, we focus on the fundamental task of OSR which is enabling classifiers to make the "none of the above" decision.

### 2.2 RF Geometry and Accuracy Preserving Proximities

Various efforts have been made to the theoretical analyses of RF in which connections between RF and other ML methods were revealed. For instance, the connection between RFs and kernel methods was highlighted in [39]. Another promising direction is exploring the connection between the nearest neighbor predictors and RFs, which was pointed out in [26] and further investigated in [4]. Inspired by this finding, in [35], the authors proposed RF-GAP and showed the proximity-weighted sum (for regression) or majority vote (for classification) using RF-GAP closely matched with the out-of-bag (OOB) prediction. This indicates RF-GAP can better capture the data geometry learned by RF, compared to other types of proximities defined for RF, e.g., the original RF proximity [8] and OOB proximity [25].

Specifically, RF-GAP proximity is a pair-wise similarity measure learned by RF from the given task. Given two observations $x_i$ and

$\mathbf{x}_j$, their RF-GAP is defined as:

$$p_{\text{GAP}}(\mathbf{x}_i, \mathbf{x}_j) = \frac{1}{|S_i|} \sum_{t \in S_i} \frac{c_j(t) \cdot I(j \in J_i(t))}{|M_i(t)|}, \quad (2)$$

where $S_i$ is the set of trees in which the observation $\mathbf{x}_i$ is OOB, $|S_i|$ denotes the size of $S_i$, $J_i(t)$ is the multiset (potentially repeated) of in-bag observations which share the terminal node with observation $\mathbf{x}_i$ in the tree $t$, $c_j(t)$ is the multiplicity of the $\mathbf{x}_j$ in the bootstrap sample, and $|M_i(t)|$ is the size of the multiset of in-bag samples in the terminal node shared with $\mathbf{x}_i$ in tree $t$ including multiplicities.

### 2.3 Gaussian Processes

GPs provide a flexible and powerful Bayesian nonparametric approach to model unknown functions and mappings which lies at the core of solving many ML tasks [40]. They have shown superior performance in many challenging tasks, for example, in patient volume forecasting [17], inflation forecasting [13], as well as causal inference [9, 10].

A GP is a stochastic process that can be seen as a (prior) distribution of a real-valued function $f(\mathbf{x})$ [50]. It is characterized by its mean function $m(\mathbf{x})$ and a covariance function $k_f(\mathbf{x}_i, \mathbf{x}_i)$, which are defined by $m(\mathbf{x}) = \mathbb{E}[f(\mathbf{x})]$, and $k_f(\mathbf{x}_i, \mathbf{x}_j) = \mathbb{E}[(f(\mathbf{x}_i) - m(\mathbf{x}_i))(f(\mathbf{x}_j) - m(\mathbf{x}_j))]$. In practice, the mean function is often set to be 0, i.e., $m(\mathbf{x}) = 0$ for every $\mathbf{x}$. When pairing with Gaussian likelihood (commonly used for GP regression) of $\mathbf{y}$, that is,

$$y = y(\mathbf{x}) = f(\mathbf{x}) + \epsilon, \quad (3)$$

where $\epsilon \sim \mathcal{N}(0, \sigma_\epsilon^2)$ is additive white Gaussian noise, we can derive the logarithm of the marginal likelihood as

$$\log p(\mathbf{y}|\mathbf{X}, \boldsymbol{\theta}) = -\frac{1}{2}\mathbf{y}^T\mathbf{K}^{-1}\mathbf{y} - \frac{1}{2}\log|\mathbf{K}| - \frac{n}{2}\log 2\pi, \quad (4)$$

where $\boldsymbol{\theta}$ denotes the hyperparameters (noise variance $\sigma_\epsilon^2$ and parameters in covariance function), $\mathbf{K} = \mathbf{K}_{ff} + \sigma_\epsilon^2\mathbf{I}$ with $\mathbf{K}_{ff}$ is the covariance matrix constructed on training data $\mathbf{X}$. The hyperparameters, i.e., $\boldsymbol{\theta}$, are learned by maximizing log-likelihood in (4) using e.g., a gradient-based optimizer.

## 3 METHODOLOGY

In this section, we describe the proposed approach in detail. Briefly, we first train a closed-set RF classifier on known classes, and construct RF-GAP proximity matrix on training samples. It is worth noting that although a notion of distance between two samples is often defined as $1 -$ proximity in the literature of OSR [35], directly using such distance (or proximity) for OSR task can be problematic. Particularly, it can be shown that points from an unbounded partition learned by RF classifier in the feature space may have exactly the same proximities to training points. To that end, in the second step, we infer a Mahalanobis distance from the RF-GAP proximity matrix using GP framework, which introduces a linear transformation to the original feature space. Finally, we utilize the linearly transformed feature space for OSR. The validity or theoretical foundation of working with linearly transformed feature space for OSR is given by [3]. Particularly, it has been shown that managing the open-set risk in the linearly transformed feature space will also manage the open-set risk in the original feature space [3].

### 3.1 Closed-set RF Classification and RF-GAP

Let $\{\mathbf{x}_i, y_i\}_{i=1}^N$ be the collection of data points for training, where $\mathbf{x}_i \in \mathbb{R}^D$ is a $D$-dimensional feature vector whose label is $y_i \in \{1, 2, \ldots C\}$. We first train a closed-set RF classifier whose decision function is $f_r(\mathbf{x}; \boldsymbol{\theta}_r)$, where $\boldsymbol{\theta}_r$ collects hyperparameters. Then we compute the corresponding RF-GAP proximity matrix $\mathbf{P}_r \in \mathbb{R}^{N \times N}$ on training data points with (2). Although it might seem counter-intuitive that the first step of enabling OSR for RF classifier is to train a closed-set RF classifier, it has been shown that the ability of a classifier to make the 'none-of-above' decision is highly correlated with its accuracy on the closed-set or known classes [47]. This can be seen from the perspective of model calibration or uncertainty quantification in which low confidence predictions are usually correlated with high error rates. More discussion on this can be found in [47]. In this work, we used scikit-learn [22] for training (multi-class) RF classifiers on known classes, the hyperparameters were selected with grid search and 5-fold cross validation (CV), the search grid is shown in Table 1.

### 3.2 Distance Metric Learning

With the trained closed-set RF classifier and its RF-GAP proximity matrix $\mathbf{P}_r$ on training samples, we can infer a distance metric, which is later used in distance-based OSR. Particularly, we learn a Mahalanobis distance that is parameterized by a positive semidefinite (PSD) matrix $\mathbf{M} \in \mathbb{R}^{D \times D}$ and it is defined as

$$d_{\mathbf{L}}(\mathbf{x}_i, \mathbf{x}_j) = \sqrt{(\mathbf{x}_i - \mathbf{x}_j)^T \mathbf{M} (\mathbf{x}_i - \mathbf{x}_j)}$$
$$= \sqrt{(\mathbf{L}\mathbf{x}_i - \mathbf{L}\mathbf{x}_j)^T (\mathbf{L}\mathbf{x}_i - \mathbf{L}\mathbf{x}_j)}, \quad (5)$$

where $\mathbf{x}_i, \mathbf{x}_j \in \mathbb{R}^D$ and $\mathbf{M} = \mathbf{L}^T\mathbf{L}$. Essentially, $\mathbf{L}\mathbf{x}$ can be seen as a linearly transformed feature space. The main idea is that using the distance metric learned from RF-GAP can improve the OSR performance, compared with Euclidean distance (in the original feature space), since label information as well as data geometry are exposed to OSR component.

We note that RF-GAP is not a proper similarity metric since it is asymmetric. This can be easily fixed by working with $\mathbf{P} = \frac{1}{2}(\mathbf{P}_r + \mathbf{P}_r^\top)$ which is symmetric, i.e., $\mathbf{P}(i, j) = \mathbf{P}(j, i)$. Given two samples, $\mathbf{x}_i$ and $\mathbf{x}_j$, we collect the Euclidean distance in each dimension between them into a vector $\mathbf{d}_{ij} = \left[d_{ij}^1, d_{ij}^2, \ldots, d_{ij}^D\right] \in \mathbb{R}^D$. Further, similar to (3), we consider the following generative model

$$\mathbf{Q}(i, j) = f(\mathbf{d}_{ij}) + \epsilon, \quad (6)$$

where $\mathbf{Q}(i, j) = 1 - \mathbf{P}(i, j)$ is the RF-GAP distance (defined as $1 -$ RF-GAP proximity) between $\mathbf{x}_i$ and $\mathbf{x}_j$, $\epsilon \sim \mathcal{N}(0, \sigma_\epsilon^2)$ is additive white noise, and the mapping $f$ is governed by a GP with zero mean, and covariance function $k_f\left(\mathbf{d}_{ij}, \mathbf{d}_{i'j'}\right)$.

**Table 1: Search grid for RF hyperparameter tuning**

| Hyperparameter | Range |
|---|---|
| Number of trees | 100-500 (step size 50) |
| Max depth | 5-50 (step size 5) and till pure leaf node |
| Max features | Sqrt, log2 of total features |
| Split criterion | Gini, Entropy |

In the GP framework, the covariance function transforms the distance between inputs to the covariance between outputs. In our work, we adopted the most widely used squared exponential covariance function (for its smoothness assumption):

$$
\begin{aligned}
k_f\left(\mathbf{d}_{ij}, \mathbf{d}_{i'j'}\right) &= \sigma_f^2 \exp(-\frac{1}{2}(\mathbf{d}_{ij} - \mathbf{d}_{i'j'})^T \mathbf{L}^T \mathbf{L}(\mathbf{d}_{ij} - \mathbf{d}_{i'j'})) \\
&= \sigma_f^2 \exp(-\frac{1}{2} d_{\mathbf{L}}^2(\mathbf{d}_{ij}, \mathbf{d}_{i'j'})),
\end{aligned}
\tag{7}
$$

where $\sigma_f^2$ is the variance of the kernel and $d_{\mathbf{L}}(\mathbf{d}_{ij}, \mathbf{d}_{i'j'}) = ||\mathbf{L}\mathbf{d}_{ij} - \mathbf{L}\mathbf{d}_{i'j'}||$ is a distance metric. To reduce the number of hyperparameters, we assume $\mathbf{L}$ is a diagonal matrix. In this case, the kernel or covariance function in (5) is known as automatic relevance determination (ARD) squared exponential kernel, since the diagonal terms in $\mathbf{L}$ (length scales) determine the model complexity in the corresponding dimensions [50]. In theory, the hyperparameters in $\theta = \{\sigma_f, \sigma_e, \mathbf{L}\}$ can be learned by maximizing (4). However, since we are computing pair-wise distance and the number of pairs scales with $N^2$, this introduces computational burden of exact inference for GP. Instead, we use sparse GP with variational inference proposed in [45], where a lower bound $\mathcal{L}$ on (4) is constructed by introducing a set of $m \ll N$ inducing points that in the same space of training samples. The learning of hyperparameters is carried out by maximizing $\mathcal{L}$. In the next section, we work with transformed feature space, i.e., $\tilde{\mathbf{x}} = \mathbf{L}\mathbf{x}$ for OSR.

In short, we infer a Mahalanobis distance from RF-GAP proximity by introducing a mapping from the Euclidean distance of each dimension in feature space to RF-GAP distance (i.e., 1 − RF-GAP proximity), as shown in (6). This mapping is governed by a GP with kernel function shown in (7) that has Mahalanobis distance inside as hyperparameters which are then optimized by maximizing the model evidence in (4).

## 3.3 Distance-based OSR

Unlike [2] and [48] in which a collection of EVT-based models were trained independently on known classes for OSR, we train a global OSR model shared across all known classes. To that end, we resort to a distance-based OSR framework built upon the nearest neighbors classifiers, which is first proposed in [29], namely open-set nearest neighbor (OSNN). Later, improved upon OSNN by extending 1-nearest neighbor to $K$-nearest neighbors (KNN) and incorporating EVT, KOSNN was proposed in [42]. Principally, both OSNN and KOSNN aim to meaningfully bound KLOS for OSR by thresholding a distance (or dis-similarity) ratio. Since KOSNN is the point of departure for the proposed model, we briefly describe here. Specifically, we use KOSNN in the transformed feature space for recognition of samples from unknown classes.

We denote a sample in the original feature space as $\mathbf{x}$, whereas its projection in the linearly transformed space as $\tilde{\mathbf{x}} = \mathbf{L}\mathbf{x}$. For a test sample $\mathbf{x}_*$ and its projection $\tilde{\mathbf{x}}_*$, we denote the set of $K$-nearest neighbors (i.e., neighborhood) as

$$
\mathcal{N}(\tilde{\mathbf{x}}_*) = \left\{ \left( \tilde{\mathbf{x}}_{(u)}, y_u \right) \middle| u = 1, \ldots, K \right\}.
$$

The label for $\mathcal{N}(\tilde{\mathbf{x}}_*)$, denoted as $y_*$, is given by majority vote, i.e.,

$$
y_* = \underset{y \in 1, \ldots, C}{\arg\max} \sum_{u=1}^{k} \mathbb{I}(y_{(u)} = y).
\tag{8}
$$

The average distance to $\tilde{\mathbf{x}}_*$ in $\mathcal{N}(\tilde{\mathbf{x}}_*)$ is denoted by $\bar{d}(\tilde{\mathbf{x}}_*)$, where

$$
\bar{d}(\tilde{\mathbf{x}}_*) = \frac{1}{K} \sum_{u=1}^{K} d\left( \tilde{\mathbf{x}}_*, \tilde{\mathbf{x}}_{(u)} \right).
\tag{9}
$$

Next, we identify another neighborhood, $\mathcal{N}^c(\tilde{\mathbf{x}}_*)$, that comprises of $K$-nearest samples around $\tilde{\mathbf{x}}_*$ that are not of class $y_*$. Similarly, the average distance to $\tilde{\mathbf{x}}$ in $\mathcal{N}^c(\tilde{\mathbf{x}}_*)$ is computed as

$$
\bar{d}^c(\tilde{\mathbf{x}}^*) = \frac{1}{K} \sum_{u=1}^{K} d\left( \tilde{\mathbf{x}}_*, \tilde{\mathbf{x}}_{(u)}^c \right).
\tag{10}
$$

The distance ratio is computed as

$$
r(\tilde{\mathbf{x}}^*) = \frac{\bar{d}(\tilde{\mathbf{x}}^*)}{\bar{d}^c(\tilde{\mathbf{x}}^*)}.
\tag{11}
$$

Let $R$ be a random variable where the distance ratios $r(\tilde{\mathbf{x}})$ are drawn samples from the distribution of $R$, and $S = R - t_R | R > t_R$ denote the excesses of $R$ above a large threshold $t_R$. By using the Pickands–Balkema–De Haan theorem [28], the probability $P(R > r(\tilde{\mathbf{x}}))$ can be approximated with

$$
\begin{aligned}
P(R > r(\tilde{\mathbf{x}})) &= P(R > t_R) \cdot P(S > r(\tilde{\mathbf{x}}) - t_R) \\
&\approx P(R > t_R) \cdot \bar{G}_{\tau, \gamma}(r(\tilde{\mathbf{x}}) - t_R),
\end{aligned}
\tag{12}
$$

where $P(R > t_R)$ can be estimated using training data, and $\bar{G}_{\tau, \gamma}(x) = (1 - \frac{x}{\tau})^{-\frac{1}{\gamma}}$ is the generalized Pareto distribution with $1 - \frac{x}{\tau} > 0$, $\gamma < 0$, and $\tau > 0$. The parameters $\tau, \gamma$ of the generalized Pareto distribution are learned by maximizing their log-likelihood.

Finally, we update the closed-set RF prediction based on the results from distance-based OSR in transformed space. The decision function of our proposed approach, namely, RF-KOSNN, to classify $\mathbf{x}_*$ is given by

$$
\hat{y}(\mathbf{x}_*) = \begin{cases} f_r(\mathbf{x}_*) & \text{if } P(R > r(\tilde{\mathbf{x}}_*)) \geq \alpha \\ c_{\text{unknown}} & \text{otherwise,} \end{cases}
\tag{13}
$$

where $f_r(\cdot)$ is decision function of the closed-set RF classifier, $\alpha$ is a probability threshold that can be interpreted as the probability that a data point from a known class is classified as unknown. The hyperparameters in KOSNN are selected using CV. In our experiments, we used 5-fold CV, and the suggested hyperparameter grid in [42].

## 4 EXPERIMENTS AND RESULTS

In this section, we present experimental results on both synthetic and real-world datasets.

### 4.1 Evaluation metrics

We employ the same sets of performance metrics in [42]. To be more specific, in testing set, data points from different unknown classes are aggregated into a class named $c_{\text{unknown}}$ (i.e., the 'none-of-the-above' class). Therefore, there are $C + 1$ classes in performance evaluation. Essentially, assessing OSR performance is evaluating multi-class classification of these $C + 1$ classes. Both precision (denoted as precisionOSR) and recall (denoted as recallOSR) on $c_{unknown}$ are used to measure performance on detecting unknown classes. We additionally compute the geometric mean between precisionOSR and recallOSR, denoted as geoMeanPR, to take the trade-off between them into account. Furthermore, to assess performance on the known classes, known class accuracy (denoted as AccKnownCLS), open-set micro-F1 (denoted as micF1) and open-set macro-F1 (denoted as macF1) are calculated. The open-set micro- and macro-F1 scores, introduced in [29], are extensions of the F1-score for the open-set multi-class setting. They are calculated by micro- or macro-averaging the precision and recall of the known classes, respectively. Finally, accuracy on all $C + 1$ classes (denoted as Accuracy) are used to characterize the overall accuracy.

**Table 2: Performance metrics on simulated data**

| | Accuracy (%) | AccKnownCLS (%) | RecallOSR (%) | PrecisionOSR (%) | geoMeanPR (%) | micF1 (%) | macF1 (%) |
|---|---|---|---|---|---|---|---|
| RF Closed-set | 35.01 | 100 | 0 | 0 | 0 | 51.87 | 59.26 |
| KOSNN | 84.17 | 96.58 | 77.49 | 97.67 | 87.00 | 81.03 | 89.89 |
| RF-KOSNN | **94.00** | 95.89 | **92.99** | 97.67 | **95.30** | **91.80** | **94.39** |
| OSNN | 67.63 | **98.63** | 50.92 | 98.57 | 70.85 | 68.09 | 85.03 |
| RF-OSNN | 84.41 | **98.63** | 76.75 | **99.05** | 87.19 | 81.59 | 90.19 |

## 4.2 Simulated Data

We consider a similar toy example in the two-dimensional plane from [42] for the convenience of visualization. Specifically, the observations are generated by sampling from 7 different multivariate Gaussian distributions. To simulate open-set scenarios, only 80% of observations from 3 clusters/classes are revealed to model in training, and the rest observations, i.e., 20% of aforementioned 3 classes as well as all observations from the other 4 classes, are reserved for testing. The training and testing sets are plotted in Fig. 1. In addition to the proposed RF-KOSNN method, for comparison purpose, we also included closed-set RF, orginal KOSNN [42], OSNN [29] and finally, OSNN equipped with the distance metric induced by RF which is denoted as RF-OSNN. For proper comparison, the same research grid was used for both RF-KOSNN and KOSNN, and similarly, we applied the same method of threshold searching described in [29] for both RF-OSNN and OSNN.

The performance metrics for these approaches are summarized in Table 2, and additionally, their corresponding confusion matrices and decision boundaries are shown in Fig. 2. Despite its perfect accuracy on known classes, the plain vanilla closed-set RF model achieved low overall accuracy and failed the OSR task, whereas other methods are able to manage the open set risk to different extents. It can be seen that RF-KOSNN achieved the best performance in terms of overall accuracy by trading minor decrease in accuracy on known classes in exchange for better performance on OSR. By adopting RF-GAP induced linear transformation, RF-KOSNN and RF-OSNN outperformed KOSNN and OSNN, respectively.

## 4.3 Open Access Real-World Datasets

The proposed approach RF-KOSNN was then tested on two popular real-world open access datasets, iris dataset and hand-written digits dataset, which are widely adopted for benchmarking. The experiments were repeated 10 times to properly accounted for randomness from both RF model and GP model, and performance metrics were averaged over all experiments. As mentioned above, KOSNN and OSNN are inherently connected, particularly, KOSNN can be seen as a generalization of OSNN, and demonstrated superior performance, therefore, here we only consider KOSNN for benchmarking.

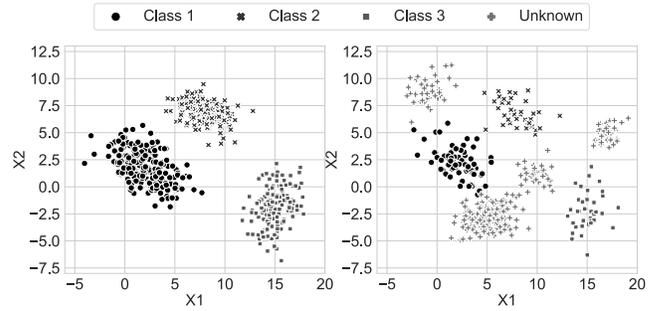

**Figure 1: Training set (left) and testing set (right) for simulated data. In training, only samples from class 1, 2 and 3 are observed, whereas in testing not only samples from class 1, 2, 3 but also samples from unknown classes (whose samples are not observed during training) are presented.**

### 4.3.1 Iris Dataset.
The iris dataset contains 150 data points from 3 different classes of iris, namely, setosa, versicolour, and virginica, where each class has 50 data points described by 4 features (sepal length, sepal width, petal length and petal width). In the experiments, we assumed that only 75% (randomly selected in each experiment) samples from setosa and virginica classes were observed in training, and the rest 25% samples from setosa and virginica along with all 50 samples from versicolour were adopted for testing. The performance metrics for both KOSNN and RF-KOSNN are shown in Fig. 3 from which we can see RF-KOSNN achieved better overall accuracy with lower variance by improving performance on identifying samples from the unknown class.

### 4.3.2 Hand-written Digits Datasets.
We used the hand-written digits datasets from scikit-learn which is a copy of the test set of the UC Irvine (UCI) ML Repository hand-written digits datasets [21]. This dataset contains 1797 instances from 10 classes corresponding to digit 0 to digit 9. The instances or data points in this datasets are images of size $8 \times 8$ (reduced from $32 \times 32$ bitmaps) with pixel intensity range from 0 to 16. More details on preprocessing procedures adopted are available in [21].

In our experiments, we randomly selected the 80% of images from the first 5 classes (from digit 0 to digit 4) for training, and the rest of (unobserved) images from the first 5 classes as well as all images from remaining classes (from digit 5 to digit 9) were adopted for testing, as illustrated in Fig. 5a. Specifically, there were 720 images in training set: 152 from class 0, 142 from class 1, 137 from class 2, 144 from class 3 and 145 from class 4. For testing set, there were 1077 images: 26 from class 0, 40 from class 1, 40 from class 2, 39 from class 3, 36 from class 4, and finally, all 896 samples from class 5 to class 9.

The experimental results are summarized in Fig. 4, and the performance metrics are averaged across all experiments. Comparing with KOSNN, RF-KOSNN achieved similar performance on classifying the known classes, and better performance on classification of the unknown classes. This indicates that RF-KOSNN is able to better manage the trade-off between empirical risk and open space risk. We further visualize the weights induced from RF-GAP in Fig.5b. It can be seen that higher weights are assigned to the dimensions (pixel locations) that are more informative in classification, for instance, location $(1, 1)$, which is more likely to fire-up only for certain classes (e.g., digit 3), whereas lower weights are assigned to the less informative locations that are commonly fired-up across all classes, e.g., location $(4, 3)$.

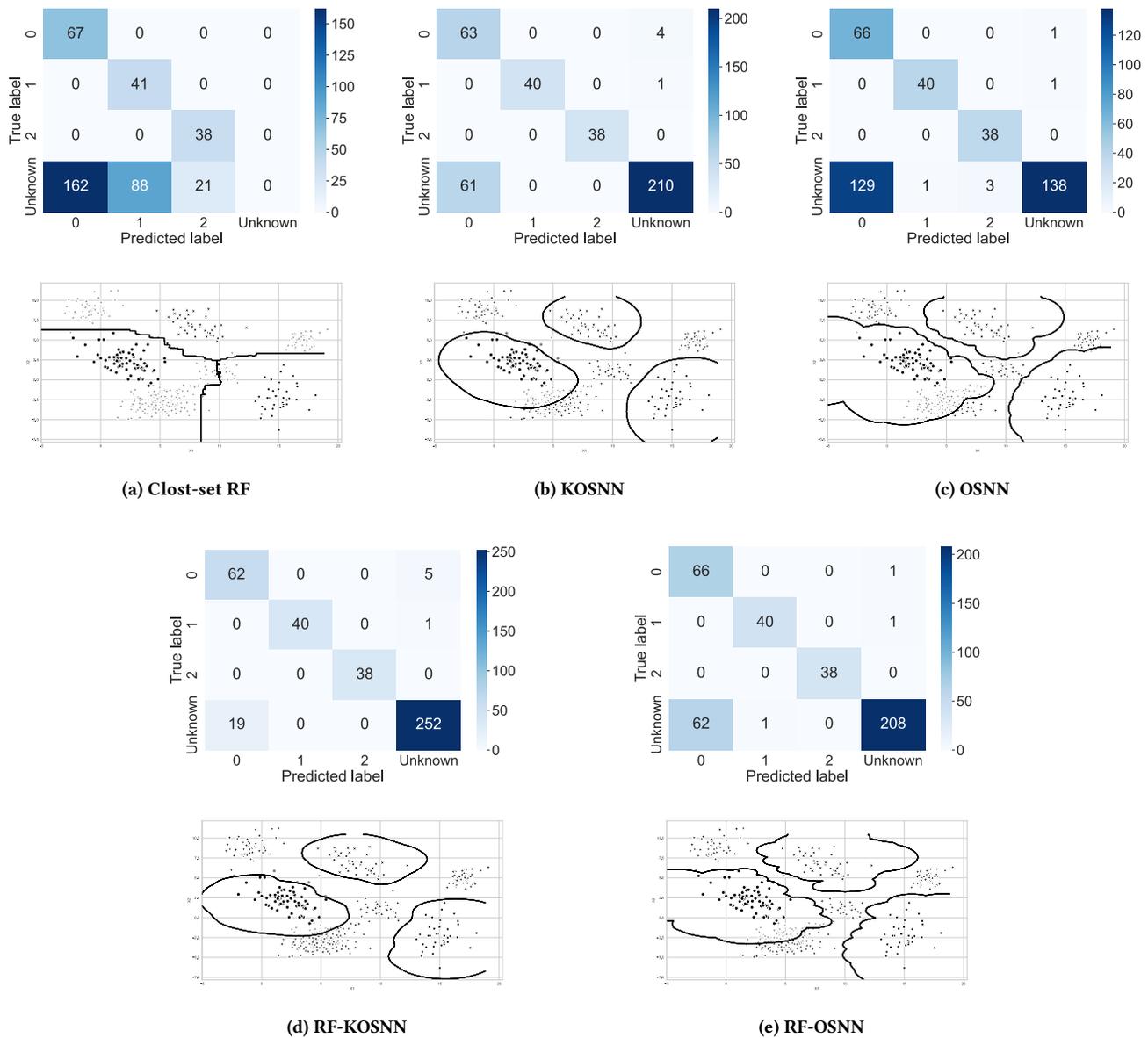

**Figure 2: Confusion matrix (top) and decision boundary on testing set (bottom) for each method on simulated data. In visualization of decision boundary, the hue of testing samples is set to classification correctness.**

## 4.4 Funds Data

Finally, we tested the proposed approach on fund data sourced from Morningstar Data Warehouse data feed in which funds are categorized into different Morningstar categories [30]. In the data, various levels of aggregation breakdowns of funds are provided based on funds' portfolio composition, e.g., breakdowns with respect to asset allocation, bond region, stock sector, and stock type, etc. with with 14 numerical and 2 categorical variables. More details regarding attributes of fund data can be found in [14].

We utilized this dataset on characteristics of funds to estimate their corresponding Morningstar categories. Particularly, in our experiments, we used the December 2022 data (which is available at beginning of January 2023) from which we selected funds in the following Morningstar categories: US Large Blend (1447 funds), US Large Growth (1278 funds) and US Large Value (1271 funds). To create the open-set scenarios, US Large Value class was considered as the unknown class, similarly, we assumed that only 80% funds (randomly selected) of US Large Blend and US Large Growth were observed for training, and the rest 20% funds in US Large Blend and US Large Growth along with 100% (all 1271 funds) for US Large Value were reserved for testing. The experiments were repeated 10 times, and corresponding performance metrics, shown Fig. 6, are averaged across 10 experiments. RF-KOSNN is able to offer better overall performance.

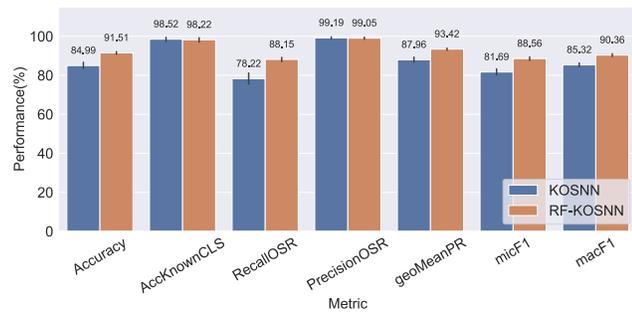

**Figure 3: Experiment results from KOSNN and RF-KOSNN on the iris data over 10 runs.**

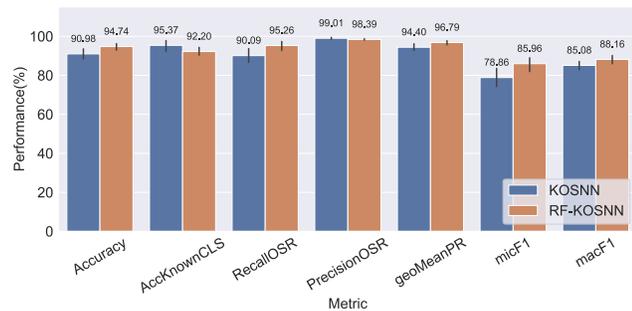

**Figure 4: Experiment results from KOSNN and RF-KOSNN on hand-written digits data over 10 runs.**

## 5 CONCLUSION

In this paper, we propose an open-set RF classifier, namely RF-KOSNN, which is built upon RF-GAP and KOSNN. Particularly, we learn a linear transformation from RF-GAP using GP framework, and we manage open-set risk in the linearly transformed feature space. Instead of training an EVT model independently for each known class, the proposed approach utilizes an EVT model that share across different known classes, which is more robust. The proposed method is tested on both synthetic and real-world datasets, where based on the experimental results, RF-KOSNN outperformed KOSNN. The proposed framework can be readily adopted for adding OSR capability for other types of tree-based approaches where proximities can be extracted.

## ACKNOWLEDGMENTS

The views expressed here are those of the authors alone and not of BlackRock, Inc.

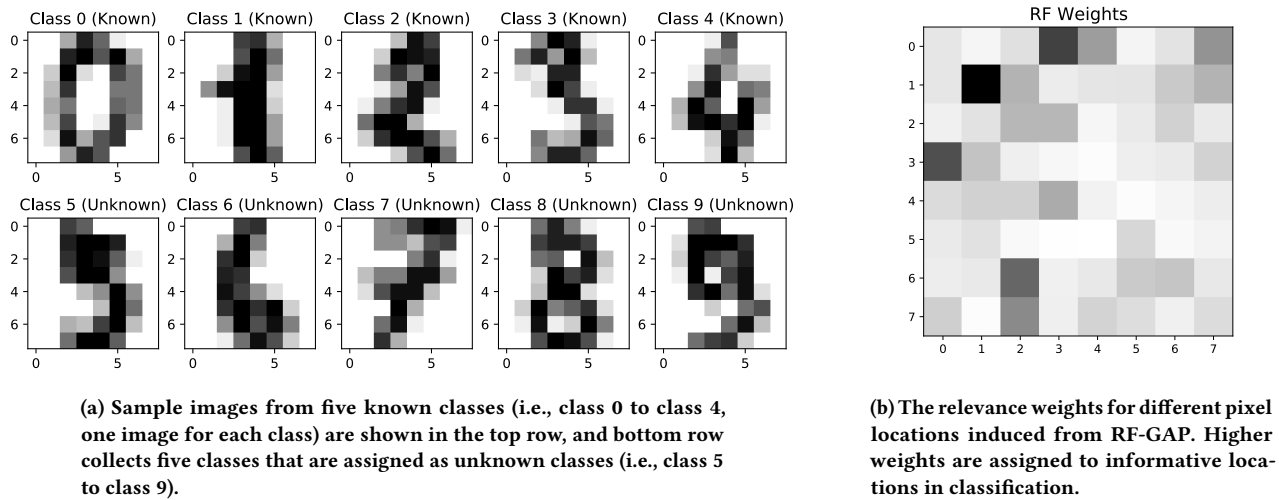

(a) Sample images from five known classes (i.e., class 0 to class 4, one image for each class) are shown in the top row, and bottom row collects five classes that are assigned as unknown classes (i.e., class 5 to class 9).

(b) The relevance weights for different pixel locations induced from RF-GAP. Higher weights are assigned to informative locations in classification.

**Figure 5: Illustration of known and unknown classes for hand-written digits datasets as well as weights induced from RF-GAP. The images are displayed in inverse gray-scale, i.e., higher values/intensities are displayed in darker gray-scale.**

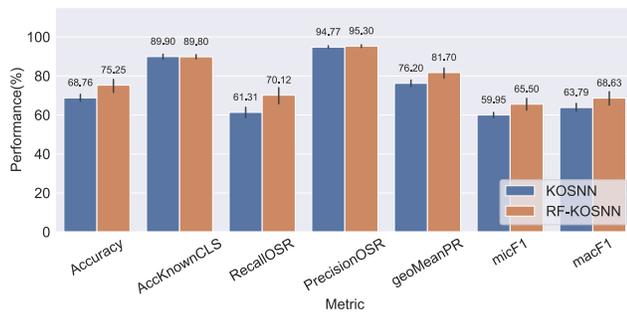

**Figure 6: Experimental results from KOSNN and RF-KOSNN on Morningstar fund data averaged over 10 runs.**